\documentclass[11pt,a4paper]{article}
\usepackage[hyperref]{emnlp2020}
\usepackage{times}
\usepackage{latexsym}

\usepackage{url}
\usepackage[T1]{fontenc}
\usepackage{array, booktabs}
\usepackage{graphicx}
\usepackage{float}
\usepackage{caption}
\usepackage{booktabs}
\usepackage{enumitem}
\usepackage{epigraph}
\usepackage[textsize=tiny]{todonotes}

\newcommand{\Level}{World Scope}
\newcommand{\level}{\textsc{WS}}

\usepackage{microtype}

\aclfinalcopy % Uncomment this line for the final submission
\def\aclpaperid{1455} %  Enter the acl Paper ID here

\title{Experience Grounds Language}

\author{
  \textnormal{Yonatan Bisk}*
  \hspace{3em} \textnormal{Ari Holtzman}* \hspace{3em} \textnormal{Jesse Thomason}* \\[5pt]
  \begin{tabular}{@{}l@{\hspace{30pt}}c@{\hspace{30pt}}c@{\hspace{30pt}}r@{}}
  Jacob Andreas & 
  Yoshua Bengio & 
  Joyce Chai & 
  Mirella Lapata
  \end{tabular}~\\
  \begin{tabular}{@{}l@{\hspace{12pt}}c@{\hspace{12pt}}c@{\hspace{12pt}}c@{\hspace{12pt}}r@{}}
  Angeliki Lazaridou & 
  Jonathan May & 
  Aleksandr Nisnevich  & 
  Nicolas Pinto &  
  Joseph Turian
  \end{tabular}
}

\date{}

\begin{document}
\maketitle

\begin{abstract}
Language understanding research is held back by a failure to relate language to the physical world it describes and to the social interactions it facilitates.
Despite the incredible effectiveness of language processing models to tackle 
tasks after being trained on text alone, successful linguistic \textit{communication} relies on a shared experience of the world.
It is this shared experience that makes utterances meaningful.

Natural language processing is a diverse field, and progress throughout its development has come from new representational theories, modeling techniques, data collection paradigms, and tasks.
We posit that the present success of representation learning approaches trained on large, text-only corpora requires 
the parallel tradition of research on the broader physical and social context of language to address the deeper questions of communication.

\end{abstract}

Improvements in hardware and data collection have galvanized progress in NLP across many benchmark tasks.
Impressive performance has been achieved in language modeling \cite{radford2019language, Zellers2019, keskar2019ctrl} and span-selection question answering \cite{devlin:naacl19, yang2019xlnet, lan2019albert} through massive data and massive models.
With models 
exceeding human performance on such tasks, now is an excellent time to reflect on a key question: 
\begin{center}\textit{Where is NLP going?}\end{center}
In this paper, we consider how the data and world a language learner is exposed to define and constrains the scope of that learner's semantics.
Meaning does not arise from the statistical distribution of words, but from their use by people to communicate.
Many of the assumptions and understandings on which communication relies lie outside of text.
We must consider what is missing from models trained solely on text corpora, even when those corpora are meticulously annotated or Internet-scale.

    \textit{You can't learn language from the radio.}
Nearly every NLP course will at some point make this claim.
The futility of learning language from linguistic signal alone is intuitive, and mirrors the belief that humans lean deeply on non-linguistic knowledge \cite{chomsky1965aspects, chomsky1980language}.
However, as a field we attempt this futility: trying to learn language from the \textit{Internet}, which stands in as the modern radio to deliver limitless language.
In this piece, we argue that the need for language to attach to ``extralinguistic events" \cite{ErvinTripp1973} and the requirement for social context \cite{Baldwin1996} should guide our research.

\begin{figure}
\setlength{\epigraphwidth}{0.9\linewidth}
\epigraph{Meaning is not a unique property of language, but a general characteristic of human activity ... We cannot say that each morpheme or word has a single or central meaning, or even that it has a continuous or coherent range of meanings ... there are two separate uses and meanings of language -- the concrete ... and the abstract.}{\nocite{Harris1954} \textit{Zellig S. Harris (Distributional Structure 1954)}}
\vspace{-20pt}
\end{figure}

Drawing inspiration from previous work in NLP, Cognitive Science, and Linguistics, we propose the notion of a \Level{} (\level{}) as a lens through which to audit progress in NLP.
We describe five \level{}s, and note that most trending work in NLP operates in the second (Internet-scale data).

We define five levels of \textbf{\Level{}}:
\begin{enumerate}[noitemsep,nolistsep]
    \item[] \level{}\ref{sec:L1}.\hspace{5pt} Corpus    \textit{(our past)}
    \item[] \level{}\ref{sec:L2}.\hspace{5pt} Internet  \textit{(most of current NLP)}
    \item[] \level{}\ref{sec:L3}.\hspace{5pt} Perception \textit{(multimodal NLP)}
    \item[] \level{}\ref{sec:L4}.\hspace{5pt} Embodiment
    \item[] \level{}\ref{sec:L5}.\hspace{5pt} Social
\end{enumerate}

These \Level{}s go beyond text to consider the contextual foundations of language: grounding, embodiment, and social interaction.
We describe a brief history and ongoing progression of how contextual information can factor into representations and tasks.
We conclude with a discussion of how this integration can move the field forward.
We believe this \Level{} framing 
serves as a roadmap for truly contextual language understanding.

\section{\level{}1: Corpora and Representations}
\label{sec:L1}
The story of data-driven language research begins with the corpus.
The Penn Treebank \cite{Marcus1993} is the canonical example of a clean subset of naturally generated language, processed and annotated for the purpose of studying representations.
Such corpora and the model representations built from them exemplify \level{}1.
Community energy was initially directed at finding formal linguistic \textit{structure}, such as recovering syntax trees. 
Recent success on downstream tasks has not required such explicitly annotated signal, leaning instead on unstructured fuzzy representations.
These representations span from dense word vectors \cite{skipgram} to contextualized pretrained representations \cite{peters-etal-2018-deep, devlin:naacl19}.

Word representations have a long history predating the recent success of deep learning methods.
Outside of NLP, philosophy \cite{austin1975things} and linguistics \cite{Lakoff1973, coleman-kay-1981} recognized that meaning is flexible yet structured.
Early experiments on neural networks trained with sequences of words \cite{Elman1990, bengio2003} suggested that vector representations could capture both syntax and semantics.
Subsequent experiments with larger models, documents, and corpora have demonstrated that representations learned from text capture a great deal of information about 
meaning in and out of context \citep{Collobert2008AUA, turian-etal-2010-word, skipgram, mccann2017learned}.

The 
intuition of such embedding representations, that context lends meaning, has long been acknowledged \cite{firth1957, turney2010frequency}.
Earlier on, discrete, hierarchical representations, such as agglomerative clustering guided by mutual information \cite{Brown1992}, were constructed with some innate interpretability.
A word's position in such a hierarchy captures semantic and syntactic distinctions.
When the Baum--Welch algorithm \cite{Welch2003} is applied to unsupervised Hidden Markov Models, it assigns a class distribution to every word, and that distribution is a partial representation of a word's ``meaning.''
If the set of classes is small, syntax-like classes are induced; if the set is large, classes become more semantic.
These representations are powerful in that they capture linguistic intuitions without supervision, but they are constrained by the structure they impose with respect to the number of classes chosen.

The intuition that meaning requires a large context, that \textit{``You shall know a word by the company it keeps."} -- \citet{firth1957}, manifested early via Latent Semantic Indexing/Analysis \cite{LSI, LSI2, dumais2004} and later in the generative framework of Latent Dirichlet Allocation \cite{LDA}.
LDA represents a document as a bag-of-words conditioned on latent topics, while
LSI/A use singular value decomposition to project a co-occurrence matrix to a low dimensional word vector that preserves locality.
These methods discard sentence structure in favor of the document.

\begin{figure}
    \includegraphics[width=\linewidth]{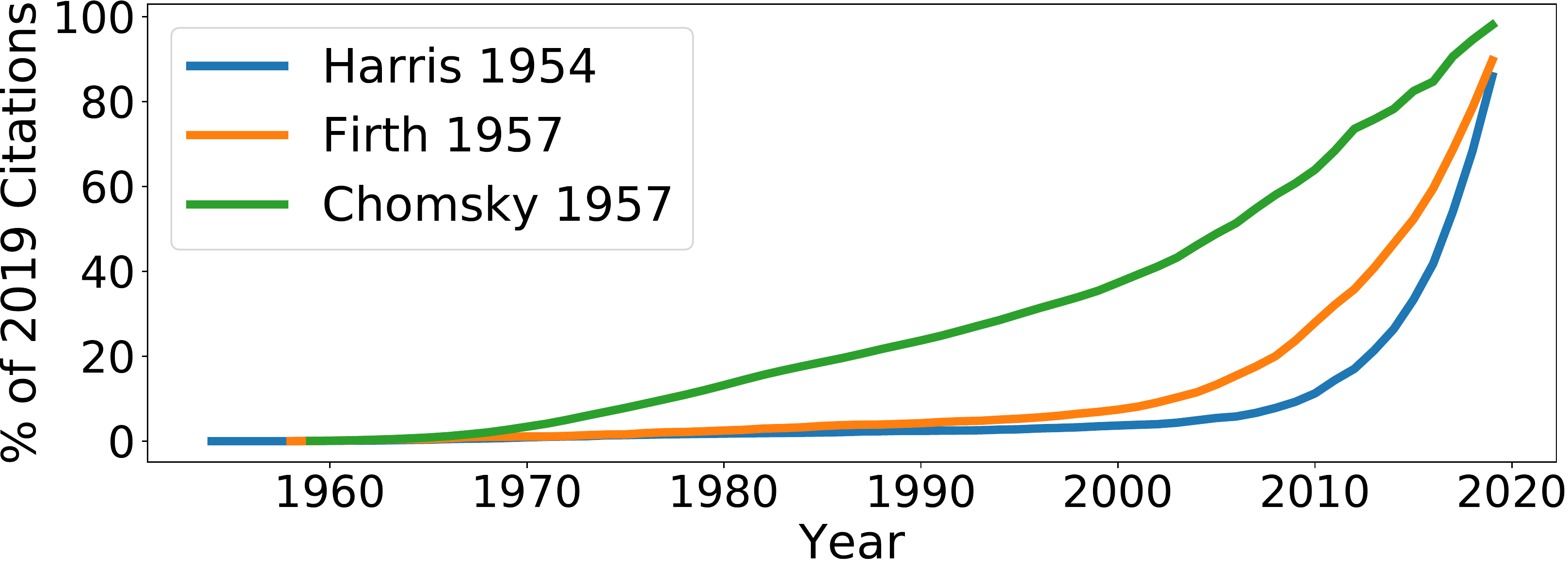}
    \begin{small}
    \textit{Academic interest in Firth and Harris increases dramatically around 2010, perhaps 
    due to the popularization of \citet{firth1957} ``You shall know a word by the company it keeps."}
    \end{small}
\end{figure}

Representing words through other words is a comfortable proposition, as it provides the illusion of definitions by implicit analogy to thesauri and related words in a dictionary definition. 
However, the recent trends in deep learning approaches to language modeling favor representing meaning in fixed-length vectors with no obvious interpretation.
The question of \textit{where} meaning resides in ``connectionist'' systems like Deep Neural Networks is an old one \cite{pollack1987,james:nips95}.
Are concepts distributed through edges or local to units in an artificial neural network?
\vspace{-3pt}
\begin{quote}
    \textit{``... there has been a long and unresolved debate between those who favor localist representations in which each processing element corresponds to a meaningful concept and those who favor distributed representations.'' \hfill \citet{HINTON19901}\\
    {\small \textit{Special Issue on Connectionist Symbol Processing}}}
\end{quote}
\vspace{-3pt}
In connectionism, words were no longer defined over interpretable dimensions or symbols, 
which were perceived as having intrinsic meaning. 
The tension of  modeling symbols and distributed representations is articulated by \citet{Smolensky1990}, and alternative representations \cite{kohonen1984,hinton-mcclelland-rumehart,barlow1989} and approaches to structure and composition \cite{erk-pado-2008-structured,Socher12Semantic} span decades of research.

The Brown Corpus \cite{browncorpus} and Penn Treebank \cite{Marcus1993} defined context and structure in NLP for decades.
Only relatively recently \cite{baroni2009wacky} has the cost of annotations decreased enough, and have large-scale web-crawls become viable, to enable the introduction of more complex text-based tasks.
This transition to larger, unstructured context (\level{}2) induced a richer semantics than was previously believed possible under the distributional hypothesis.
\section{\level{}2: The Written World}
\label{sec:L2}
Corpora in NLP have broadened to include large web-crawls.
The use of unstructured, unlabeled, multi-domain, and multilingual data broadens our world scope, in the limit, to everything humanity has ever written.\footnote{A parallel discussion would focus on the hardware required to enable advances to higher \Level{}s.
Playstations \cite{pinto2009high} and then GPUs \cite{NIPS2012_4824} made many WS2 advances possible.
Perception, interaction, and robotics leverage other new hardware.}
We are no longer constrained to a single author or source, and the temptation for NLP is to believe everything that needs knowing can be learned from the written world. 
But, a large and noisy text corpus is still a text corpus.

This move towards using large scale raw data has led to substantial advances in performance on existing and novel community benchmarks \cite{devlin:naacl19, brown2020fewshot}. 
Scale in data and modeling has demonstrated that a single representation can discover both rich syntax and semantics without our help
\cite{tenney-etal-2019-bert}.  
This change is perhaps best seen in transfer learning enabled by representations in deep models.
Traditionally, transfer learning relied on 
\textit{our} understanding of model classes, such as English grammar.
Domain adaptation simply required sufficient data to capture lexical variation, by
assuming most higher-level structure would remain the same.
Unsupervised representations 
today capture deep associations across multiple domains, and can be used successfully transfer knowledge into surprisingly diverse contexts \cite{brown2020fewshot}. 

These representations require scale in terms of both data and parameters.
Concretely, \citet{skipgram} trained on 1.6 billion tokens, while \citet{pennington-etal-2014-glove} scaled up to 840 billion tokens from Common Crawl.
Recent approaches have made progress by substantially increasing the number of model parameters to better consume these vast quantities of data. 
Where \citet{peters-etal-2018-deep} introduced ELMo with $\sim\!\!10^8$ parameters, Transformer models \cite{Vaswani2017} have continued to scale by orders of magnitude between papers \cite{devlin:naacl19,radford2019language,Zellers2019} to $\sim\!\!10^{11}$ \cite{brown2020fewshot}.

Current models are the next (impressive) step in language modeling which started with \citet{Good1953}, the weights of \citet{Kneser1995,Chen1996}, and the power-law distributions of \citet{Teh2006a}.
Modern approaches to learning dense representations allow us to better estimate these distributions from massive corpora. 
However, modeling lexical co-occurrence, no matter the scale, is still modeling the \textit{written} world.
Models constructed this way blindly search for symbolic co-occurences void of meaning.

How can models 
yield both ``impressive results'' and ``diminishing returns''?  Language modeling---the modern workhorse of neural NLP systems---is a canonical example. Recent pretraining literature has produced results that few could have predicted, crowding leaderboards with 
``super-human" accuracy \cite{rajpurkar2018know}. However, there are diminishing returns. For example, on the LAMBADA dataset \cite{paperno2016lambada}, designed to capture human intuition, GPT2 \cite{radford2019language} (1.5B), Megatron-LM \cite{shoeybi2019megatron} (8.3B), and TuringNLG \cite{rosset_2020} (17B) perform within a few points of each other and very far from perfect (<68\%). When adding another \textit{order of magnitude} of parameters (175B) \citet{brown2020fewshot} gain 
8 percentage-points, impressive but still leaving 25\% unsolved.
Continuing to expand hardware, data sizes, and financial compute cost by orders of magnitude will yield further gains, but the slope of the increase is quickly decreasing.

The aforementioned approaches for learning transferable representations 
demonstrate that sentence and document context provide powerful signals for learning aspects of meaning, especially semantic relations among words \citep{fu2014learning} and inferential relationships among sentences \cite{wang2018glue}.
The extent to which they capture deeper notions of contextual meaning remains an open question.
Past work has found that pretrained word and sentence 
representations fail to capture many grounded features of words \citep{lucy2017distributional} and sentences, 
and current NLU systems fail on the thick tail of experience-informed inferences, such as hard coreference problems \cite{peng2015solving}. ``I parked my car in the compact parking space because it looked (big/small) enough.'' still presents problems for text-only learners.

As text pretraining schemes seem to be reaching 
the point of diminishing returns, 
even for some 
syntactic phenomena \cite{vanschijndel2019quantity}, we posit that other forms of supervision, such as multimodal perception \cite{ilharco-etal-2019-large}, are 
necessary to learn the remaining aspects of meaning in context.  Learning by observation should not be a purely linguistic process, since leveraging and combining the patterns of multimodal perception can combinatorially boost the amount of signal in data through cross-referencing and synthesis.

\section{\level{}3: The World of Sights and Sounds}
\label{sec:L3}
Language learning needs perception, because perception forms the basis for many of our semantic axioms.
Learned, physical heuristics, such as the fact that a falling cat will land quietly, are generalized and abstracted into language metaphors like \textit{as nimble as a cat}~\cite{lakoff1980}.
World knowledge forms the basis for how people make entailment and reasoning decisions, commonly driven by mental simulation and analogy \cite{hofstadter2013surfaces}. 
Perception is the foremost source of reporting bias.  
The assumption that we all see and hear the same things informs not just what we name, but what we choose to assume and leave unwritten.
Further, there exists strong evidence that children require grounded sensory perception, not just speech, to learn language \cite{sachs1981language,ogrady2005,vigliocco2014language}.

Perception includes auditory, tactile, and visual input.
Even restricted to purely linguistic signals, sarcasm, stress, and meaning can be implied through prosody.
Further, tactile senses lend meaning, both physical~\cite{sinapov:icra14,Thomason2016} and abstract, to concepts like \emph{heavy} and \emph{soft}.
Visual perception is a rich signal for modeling a vastness of experiences in the world that cannot be documented by text alone~\cite{harnad:phys90}.

For example, frames and scripts \cite{schank+abelson77,Charniak1977,Dejong1981,Mooney1985} require understanding often unstated sets of pre- and post-conditions about the world.
To borrow from \citet{Charniak1977}, how should we learn the meaning, method, and implications of \textit{painting}?
A web crawl of knowledge from an exponential number of possible how-to, text-only guides and manuals \cite{Bisk2020} is misdirected without \textit{some} fundamental referents to which to ground symbols.
Models must be able to watch and recognize objects, people, and activities to understand the language describing them~\cite{hake, krishnavisualgenome,yatskar2016,perlis:aaai16} and access fine-grained notions of causality, physics, and social interactions.

While the NLP community has played an important role in the history of 
grounding \cite{mooney:aaai08}, recently remarkable 
progress 
has taken place in the Computer Vision community.
\begin{figure}
    \includegraphics[width=\linewidth]{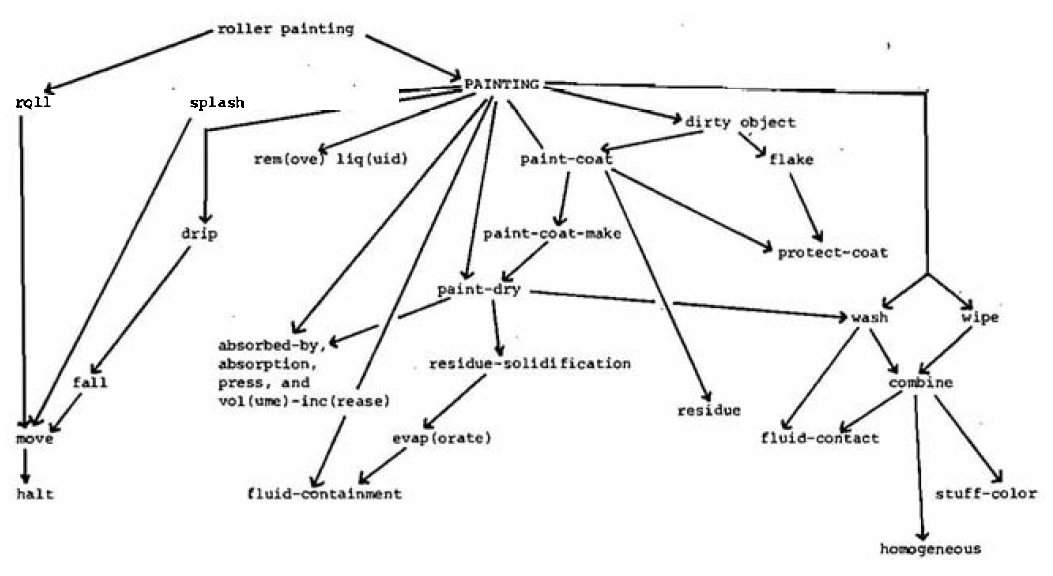}\\
    \begin{small}\textit{Eugene Charniak (A Framed PAINTING: The Representation of a Common Sense Knowledge Fragment 1977)}\end{small}
\end{figure}
It is tempting to assume that vision models trained to identify 1,000 ImageNet classes \cite{ILSVRC15}\footnote{Or the 1,600 classes of \citet{Anderson2017}. 
} are limited to extracting a bag of visual words.
In reality, Computer Vision has been making in-roads into complex visual, physical, and social phenomena, while providing reusable infrastructure.\footnote{Torchvision/Detectron2 include dozens of trained models.} 
The stability of these architectures allows for new research into more challenging world modeling.
\citet{Mottaghi2016} predicts the effects of forces on objects in images. 
\citet{Bakhtin2019} extends this physical reasoning to complex puzzles of cause and effect.
\citet{Sun2019-video, Sun2019b} models scripts and actions, and 
alternative unsupervised training regimes \cite{bachman2019amdim} open up research towards automatic concept formation.

Advances in computer vision 
have enabled building semantic representations rich enough to interact with natural language.
In the last decade of work descendant from image captioning \cite{Farhadi2010,daume12midge}, a myriad of tasks on visual question answering \cite{Antol2015,das2018embodied,yagcioglu-etal-2018-recipeqa}, natural language and visual reasoning \cite{suhr-etal-2019-corpus}, visual commonsense \cite{Zellers2019a}, and multilingual captioning/translation via video \cite{Wang_2019_ICCV} have emerged.
These combined text and vision benchmarks are rich enough to train large-scale, multimodal transformers \cite{Li2019,Lu2019,Zhou2019a} without language pretraining (e.g. via conceptual captions \cite{sharma-etal-2018-conceptual}) or further broadened to include audio \cite{tsai-etal-2019-multimodal}. Vision can also help ground speech signals \cite{Tejas2020,harwath2019} to facilitate discovery of linguistic concepts \cite{harwath2019learning}.

At the same time, NLP resources contributed to the success of these vision backbones.
Hierarchical semantic representations emerge from ImageNet classification pretraining partially due to class hypernyms owed to that dataset's WordNet origins.
For example, the \textit{person} class sub-divides into many professions and hobbies, like \textit{firefighter}, \textit{gymnast}, and \textit{doctor}.
To differentiate such sibling classes, learned vectors can also encode lower-level characteristics like clothing, hair, and typical surrounding scenes. 
These representations allow for pixel level masks and skeletal modeling, and can be extended to zero-shot settings targeting all 20K ImageNet categories \cite{chaozeroshot,changpinyozeroshot}.
Modern architectures also 
learn to differentiate instances within a general class, such as \textit{face}.
For example, facial recognition benchmarks require distinguishing over 10K unique faces \cite{liu2015faceattributes}.
While vision is by no means ``solved,'' benchmarks have led to off-the-shelf tools for building representations rich enough to identify tens of thousands of objects, scenes, and individuals. 

A \level{}3 agent, having access to potentially endless hours of video data showing the intricate details of daily comings and goings, procedures, and events, reduces susceptibility to the reporting bias of \level{}2. An ideal \level{}3 agent will exhibit better long-tail generalization and understanding than any language-only system could.
This generalization should manifest in existing benchmarks, but would be most prominent in a test of zero-shot circumstances, such as ``Will \textit{this} car fit through \textit{that} tunnel?,'' and rarely documented behaviors as examined in script learning.
Yet the \level{}3 agent will likely fail to answer, "Would a ceramic or paper plate make a better frisbee?"
The agent has not tried to throw various objects and understand how their velocity and shape interact with the atmosphere to create lift.
The agent cannot test novel hypotheses by intervention and action in the world.

\begin{figure}
\setlength{\epigraphwidth}{0.9\linewidth}
\epigraph{If A and B have some environments in common and some not ...  we say that they have different meanings, the amount of meaning difference corresponding roughly to the amount of difference in their environments ...}{\textit{Zellig S. Harris (Distributional Structure 1954)}}
\end{figure}

\section{\level{}4: Embodiment and Action}
\label{sec:L4}
In human development, \textit{interactive} multimodal sensory experience forms the basis of action-oriented categories~\cite{thelen:mit96} as children learn how to manipulate their perception by manipulating their environment.
Language grounding enables an agent to connect words to these action-oriented categories for communication~\cite{smith:mit05}, but requires action to fully discover such connections.
Embodiment---situated action taking---is therefore a natural next broader context.

An embodied agent, whether in a virtual world, such as a 2D Maze \cite{macmahon:aaai06}, a grid world~\citep{chevalier+al-ICLR2019}, a simulated house \cite{Anderson2017a,Thomason2019a,shridhar:cvpr2020}, or the real world \cite{Tellex2011,matuszek2018IJCAI,thomason:jair20,tellex:arcras:20} must translate from language to action.
Control and action taking open several new dimensions to understanding and actively learning about the world.
Queries can be resolved via dialog-based exploration with a human interlocutor~\cite{liu:aaai15}, even as new object properties, like texture and weight~\cite{thomason:corl17}, or feedback, like muscle activations~\cite{moro-kennington-2018-multimodal}, become available.
We see the need for embodied language with complex meaning when thinking deeply about even the most innocuous of questions:

\begin{quote}
    \textit{Is an orange more like a baseball or more like a banana?}
\end{quote}

\level{}1 is likely not to have an answer beyond that the objects are common nouns that can both be held.
\level{}2 may capture that oranges and baseballs both roll, but is not the deformation strength, surface texture, or relative sizes of these objects \cite{elazar2019large}.
\level{}3 may realize the relative deformability of these objects, but is likely to confuse how much force is necessary given that baseballs are used much more roughly than oranges.
\level{}4 can appreciate the nuances of the question---the orange and baseball afford similar manipulation \textit{because} they have similar texture and weight, while the orange and banana both contain peels, deform, and are edible.
People can reason over rich representations of common objects that these words evoke.

\textit{Planning} is where people first learn abstraction and simple examples of post-conditions through trial and error.
The most basic scripts humans learn start with moving our own bodies and achieving simple goals as children, such as stacking blocks.
In this space, we have unlimited supervision from the environment and can learn to generalize across plans and actions.
In general, simple worlds do not entail simple concepts: even in a block world concepts like ``mirroring'' appear~\cite{Bisk2018}.
Humans generalize and apply physical phenomena to abstract concepts with ease.

In addition to learning basic physical properties of the world from interaction, \level{}4 also allows the agent to construct rich pre-linguistic representations from which to generalize.
\citet{hespos2004} show pre-linguistic category formation within children that are then later codified by social constructs.
Mounting evidence seems to indicate that children have trouble transferring knowledge from the 2D world of books \cite{bar2013} and iPads \cite{Lin2017} to the physical 3D world.
So while we might choose to believe that we can encode parameters \cite{chomsky1981} more effectively and efficiently than evolution provided us, developmental experiments indicate doing so without 3D interaction 
may prove difficult.

Part of the problem is that much of the knowledge humans hold about the world is intuitive, possibly incommunicable by language, but still required to understand language. Much of this knowledge revolves around physical realities that real-world agents will encounter.
Consider how many explicit and implicit metaphors are based on the idea that far-away things have little influence on manipulating local space: ``a distant concern'' and ``we'll cross that bridge when we come to it.''

Robotics and embodiment are not available in the same off-the-shelf manner as computer vision models.
However, there is rapid progress in simulators and commercial robotics, and as language researchers we should match these advances at every step.
As action spaces grow, we can study complex language instructions in simulated homes \cite{shridhar:cvpr2020} or map language to physical robot control \cite{blukis2019,chai:ijcai18}. 
The last few years have seen massive advances in both high fidelity simulators for robotics \cite{todorov2012mujoco,pybullet,isaac,xiang:cvpr20} and the cost and availability of commodity hardware \cite{baxter,campeau2019kinova,murali2019pyrobot}.

As computers transition from desktops to pervasive mobile and edge devices, we must make and meet the expectation that NLP can be deployed in any of these contexts.  Current representations have very limited utility in even the most basic robotic settings \cite{Scalise2019}, making collaborative robotics \cite{rosenthal2010effective} largely a domain of custom engineering rather than science.

\section{\level{}5: The Social World}
\label{sec:L5}

Interpersonal communication is the foundational use case of natural language \cite{dunbar1993coevolution}. The physical world gives meaning to metaphors and instructions, but utterances come from a source with a purpose. Take J.L. Austin's classic example of ``BULL'' being written on the side of a fence in a large field \cite{austin1975things}.
It is a fundamentally \textit{social} inference to realize that this word indicates the presence of a dangerous creature, and that the word is written on the \textit{opposite side} of the fence from where that creature lives. 

Interpersonal dialogue as a grand test for AI is older than the term ``artificial intelligence,'' beginning at least with \citet{turing2009computing}'s Imitation Game. Turing was careful to show how easily a na{\"i}ve tester could be tricked. 
Framing, such as suggesting that a chatbot speaks English as a second language \cite{sample2014scientists}, 
can create the appearance of genuine content where there is none~\cite{weizenbaum1966eliza}.
This phenomenon 
has been noted countless times, from criticisms of Speech Recognition as ``deceit and glamour''~\cite{pierce1969whither} to complaints of humanity's ``gullibility gap''~\cite{marcus2019rebooting}. 
We instead focus on why the social world is vital to \textit{language learning}.

\begin{figure}
\setlength{\epigraphwidth}{0.9\linewidth}
\epigraph{In order to talk about concepts, we must understand the importance of mental models... we set up a model of the world which serves as a framework in which to organize our thoughts.
We abstract the presence of particular objects, having properties, and entering into events and relationships.}
{\textit{Terry \citeauthor{Winograd1971} - 1971}}
\end{figure}

\paragraph{Language that Does Something} Work in the philosophy of language has long suggested that \textit{function} is the source of meaning, as famously illustrated through Wittgenstein's ``language games'' \cite{wittgenstein1953philosophical, wittgenstein1958blue}.
In linguistics, the usage-based theory of language acquisition suggests that constructions that are useful are the building blocks for everything else \cite{langacker1987foundations,langacker1991foundations}. The \textit{economy} of this notion of use has been the subject of much inquiry and debate \cite{grice1975logic}.
In recent years, these threads have begun to shed light on what use-cases language presents in both acquisition and its initial origins in our species \cite{tomasello2009constructing, barsalou2008grounded}, indicating the fundamental role of the social world.

\level{}1, \level{}2, \level{}3, and \level{}4 
expand the factorizations of information available to linguistic meaning.
allows language to be a \textit{cause} instead of just a source of data.
This is the ultimate goal for a 
language learner: to generate language that \textit{does} something to the world.

Passive creation and evaluation of generated language separates generated utterances from their effects on other people, and while the latter is a rich learning signal it is inherently difficult to annotate.
In order to learn the effects language has on the world, an agent must \textit{participate} in linguistic activity, such as negotiation \cite{yang-etal-2019-lets, he2018decoupling, lewis2017deal}, collaboration~\cite{Chai_Fang_Liu_She_2017}, visual disambiguation \cite{Anderson2017a, lazaridou2016multi, liu:aaai15}, or providing emotional support \cite{rashkin-etal-2019-towards}.
These activities require inferring mental states and social outcomes---a key area of interest in itself~\cite{Zadeh_2019_CVPR}. 

What ``lame'' means in terms of discriminative information is always at question:
it can be defined as ``undesirable,'' but what it tells one about the processes operating in the environment requires social context to determine \cite{bloom2002children}.
It is the toddler's social experimentation with ``You're so lame!'' that gives the word weight and definite intent \cite{ornaghi2011role}.
In other words, the discriminative signal for the most foundational part of a word's meaning can only be observed by its effect on the world, and active experimentation is key to learning that effect.
Active experimentation with language starkly contrasts with the disembodied chat bots that are the focus of the current dialogue community \cite{roller:arxiv20, adiwardana2020towards, zhou2020design, chen2018gunrock, serban2017deep}, which often do not learn from individual experiences and whose environments are not persistent enough to learn the effects of actions.

\paragraph{Theory of Mind}
When attempting to get what we want, we confront people who have their own desires and identities.
The ability to consider the feelings and knowledge of others is now commonly referred to as the ``Theory of Mind'' \cite{nematzadeh2018evaluating}.
This paradigm has also been described under the ``Speaker-Listener'' model \cite{stephens2010speaker}, and a rich theory to describe this computationally is being actively developed under the Rational Speech Act Model \cite{frank2012predicting, bergen2016pragmatic}.

A series of challenges that attempt to address this fundamental aspect of communication have been introduced \cite{nematzadeh2018evaluating,sap-etal-2019-social}.
These works are a great start towards deeper understanding, but static datasets can be problematic due to the risk of 
embedding spurious patterns and bias \cite{devries:arxiv20, le-etal-2019-revisiting, gururangan2018annotation, glockner2018breaking}, especially because examples where annotators cannot agree (which are usually thrown out before the dataset is released) still occur in real use cases. More flexible, dynamic evaluation \cite{zellers2020evaluating, dinan2019build} are a partial solution, but true persistence of identity and adaption to change are both necessary and still a long way off.

Training data in \level{}1-4, complex and large as it can be, does not offer the discriminatory signals that make the hypothesizing of consistent identity or mental states an efficient path towards lowering perplexity or raising accuracy~\cite{liu2016not,devault2006societal}.
First, there is a lack of inductive bias \cite{martin2018event}. Models learn what they need to discriminate between potential labels, and it is unlikely that universal function approximators such as neural networks would ever reliably posit that people, events, and causality exist without being biased towards such solutions  \cite{mitchell1980need}.
Second, current cross entropy training losses actively discourage learning the tail of the distribution properly, as statistically infrequent events are drowned out \cite{pennington-etal-2014-glove, holtzman2019curious}.
Meanwhile, it is precisely human's ability to draw on past experience and make zero-shot decisions that AI aims to emulate.

\paragraph{Language in a Social Context}
Whenever language is used between 
people, it exists in a concrete social context: status, role, intention, and countless other variables intersect at a specific point \cite{wardhaugh2011introduction}.
These complexities are overlooked through selecting labels on which crowd workers agree.
Current notions of ground truth in dataset construction are based on crowd consensus bereft of social context.
We posit that ecologically valid evaluation of generative models will require the construction of situations where artificial agents are considered to have enough identity to be granted \textit{social standing} for these interactions.

Social interaction is a precious signal, but initial studies have been strained by the training-validation-test set scenario and reference-backed evaluations. Collecting data
about rich natural situations is often impossible.
To address this gap, learning by participation, where users can freely interact with an agent, is a necessary step to the ultimately \textit{social} venture of communication.
By exhibiting different attributes and sending varying signals, the sociolinguistic construction of identity \cite{ochs1993constructing} could be examined more deeply.
Such experimentation in social intelligence is simply not possible with a fixed corpus.
Once models are expected to be interacted with when tested, probing their decision boundaries for simplifications of reality and a lack of commonsense knowledge as in \citeauthor{Gardner:2020, Kaushik2020Learning} will become natural.

\section{Self-Evaluation}
 
We use the notion of \Level{}s to make the following concrete claims:
\subsection*{You can't learn language ...}

\paragraph{... from the radio (Internet).  \hfill \level{}2 $\subset$ \level{}3}\mbox{}
\begin{quote}
\textit{A task learner cannot be said to be in \level{}3 if it can succeed without perception (e.g., visual, auditory).}
\end{quote}

\paragraph{... from a television. \hfill \level{}3 $\subset$ \level{}4}\mbox{}
\begin{quote}
\textit{A task learner cannot be said to be in \level{}4 if the space of its world actions and consequences can be enumerated.}
\end{quote}

\paragraph{... by yourself. \hfill \level{}4 $\subset$ \level{}5}\mbox{}
\begin{quote}
    \textit{A task learner cannot be said to be in \level{}5 unless achieving its goals requires cooperating with a human in the loop.}
\end{quote}

By these definitions, most of NLP research still resides in \level{}2.
This fact does not invalidate the utility or need for any of the research within NLP, but it is to say that much of that existing research targets a different goal than \textit{language learning}.

\begin{figure}
\epigraph{These problems include the need to bring meaning and reasoning into systems that perform natural language processing, the need to infer and represent causality, the need to develop computationally-tractable representations of uncertainty and the need to develop systems that formulate and pursue long-term goals.}{\textit{\nocite{jordan2019artificial} Michael Jordan (Artificial intelligence -- the revolution hasn't happened yet, 2019)}}
\vspace{-10pt}
\end{figure}

\paragraph{Where Should We Start?} 
Many in our community are already examining phenomena in \level{}s 3-5.
Note that research can explore higher \level{} phenomena without a resultant learner being in a higher \level{}.
For example, a chatbot can investigate principles of the social world, but still lack the underlying social standing required for \level{}5. Next we describe four language use contexts which we believe are both research questions to be tackled and help illustrate the need to move beyond \level{}2.

\textbf{Second language acquisition} when visiting a foreign country leverages a shared, social world model that allows pointing to referent objects and miming internal states like hunger.
The interlingua is physical and experiential.
Such a rich internal world model should also be the goal for MT models: starting with images \cite{huang-etal-2020-unsupervised}, moving through simulation, and then to the real world.

\textbf{Coreference and WSD} leverage a shared scene and theory of mind. To what extent are current coreference resolution issues resolved if an agent models the listener's desires and experiences explicitly rather than looking solely for adjacent lexical items?
This setting is easiest to explore in embodied environments, but is not exclusive to them (e.g., TextWorld \cite{cote18textworld}).

\textbf{Novel word learning} from tactile knowledge and use: What is the instrument that you wear like a guitar but play like a piano?
Objects can be described with both gestures and words about appearance and function.
Such knowledge could begin to tackle physical metaphors that current NLP systems struggle with.

\textbf{Personally charged language:}
How should a dialogue agent learn what is hurtful to a specific person? To someone who is sensitive about their grades because they had a period of struggle in school, the sentiment of ``Don't be a fool!'' can be hurtful, while for others it may seem playful.
Social knowledge is requisite for realistic understanding of sentiment in situated human contexts.

\paragraph{Relevant recent work}
The move from \level{}2 to \level{}3 
requires 
rethinking existing tasks and investigating where their semantics can be expanded and grounded. This idea is not new \cite{chen:icml08, feng-lapata-2010-topic, bruni2014, lazaridou-etal-2016-red} and has accelerated in the last few years. \citet{multi30k} reframes machine translation with visual observations, a trend extended into videos \cite{Wang_2019_ICCV}. 
\citet{tacos:regnerietal:tacl} introduce a foundational dataset aligning text descriptions and semantic annotations of actions with videos. 
Vision can even inform core tasks like syntax \cite{shi-etal-2019-visually} and language modeling \cite{ororbia-etal-2019-like}.
Careful design is key, as visually augmented tasks can fail to require sensory perception~\cite{thomason:naacl19}.

Language-guided, embodied agents invoke many of the challenges of \level{}4.
Language-based navigation~\cite{Anderson2017a} and task completion~\cite{shridhar:cvpr2020} in simulation environments ground language to actions, but even complex simulation action spaces can be discretized and enumerated.
Real world, language-guided robots for task completion~\cite{tellex:rss14} and learning~\cite{she:sigdial14} face challenging, continuous perception and control~\cite{tellex:arcras:20}.
Consequently, research in this space is often restricted to small grammars~\cite{paul:ijrr18,walter:rss13} or controlled dialog responses~\cite{thomason:jair20}.
These efforts to translate language instructions to actions build towards using language for end-to-end, continuous control (\level{}4).

Collaborative games have long served as a testbed for studying language \cite{Werner1991} and emergent communication~\cite{schlangen2019grounded, ref-games-2018, chaabouni2020compositionality}.
\citet{suhr-etal-2019-executing} introduced an environment for evaluating language understanding in the service of a shared goal, and \citet{andreas-klein-2016-reasoning} use a 
visual paradigm for studying pragmatics.
Such efforts help us examine how inductive biases and environmental pressures build towards socialization (\level{}5), even if full social context is still too difficult and expensive to be practical.
 
Most of these works provide resources such as data, code, simulators and methodology for evaluating the multimodal content of linguistic representations \cite{schlangen2019language,silberer-lapata-2014-learning,bruni-etal-2012-distributional}.
Moving forward, we 
encourage a broad re-examination of how NLP 
frames the relationship between meaning and context~\cite{benderclimbing} and how pretraining obfuscates our ability to measure generalization~\cite{linzencan}.

\section{Conclusions}
Our 
\Level{}s are steep steps. 
\level{}5 implies a persistent agent experiencing time and a personalized set of experiences.
With few exceptions \cite{carlson2010toward}, machine learning models have been confined to IID datasets that lack 
the structure in time from which humans draw correlations about long-range causal dependencies. 
What if a machine was allowed to participate consistently?
This is difficult to test 
under current evaluation paradigms for generalization.
Yet, this is the structure of generalization in human development: drawing analogies to episodic memories and gathering new data through non-independent experiments.

As with many who have analyzed the history of NLP, 
its trends \cite{church2007}, its maturation toward a science \cite{steedman-2008-last}, and its major challenges \cite{Hirschberg2015,McClelland:2019}, we hope to provide momentum for a direction many are already heading.
We call for and embrace the incremental, but purposeful, contextualization of language in human experience.
With all that we have learned about what words can tell us and what they keep implicit, now is the time to ask: \emph{What tasks, representations, and inductive-biases will fill the gaps?}

Computer vision and speech recognition are mature enough for 
investigation of broader linguistic contexts (\level{}\ref{sec:L3}).
The robotics industry is rapidly developing commodity hardware and sophisticated software that both facilitate new research and expect to incorporate language technologies (\level{}\ref{sec:L4}).
Simulators and videogames provide potential environments for social language learners (\level{}\ref{sec:L5}).
Our call to action is to encourage the community to lean in to trends prioritizing grounding and agency, and explicitly aim to broaden the corresponding \Level{}s available to our models.

\vspace{10pt}
\section*{Acknowledgements}
Thanks to Raymond Mooney for suggestions, Paul Smolensky for disagreements, Catriona Silvey for developmental psychology help, and to a superset of: 
Emily Bender,
Ryan Cotterel,
Jesse Dunietz,
Edward Grefenstette,
Dirk Hovy,
Casey Kennington,
Ajay Divakaran,
David Schlangend,
Diyi Yang,
and Semih Yagcioglu
for pointers and suggestions.

\bibliography{main}
\bibliographystyle{acl_natbib}

\end{document}